\documentclass[letterpaper]{article}
\usepackage{aaai}
\usepackage{times}
\usepackage{helvet}
\usepackage{courier}
\usepackage{amsmath}
\usepackage{graphicx}
\usepackage{amssymb}
\usepackage{booktabs}
\usepackage[export]{adjustbox}

\DeclareMathOperator*{\argmaxB}{argmax}

\begin{document}
 \bibliographystyle{aaai}
%
\title{MACOptions: Multi-Agent Learning with Centralized Controller and Options Framework}
\author{Alakh Aggarwal*, Rishita Bansal*, Parth Padalkar*, Sriraam Natarajan\\
The University of Texas at Dallas\\
}
\maketitle
\begin{abstract}
\begin{quote}
These days automation is being applied everywhere. In every environment, planning for the actions to be taken by the agents is an important aspect. In this paper, we plan to implement planning for multi-agents with a centralized controller. We compare three approaches: random policy, Q-learning, and Q-learning with Options Framework. We also show the effectiveness of planners by showing performance comparison between Q-Learning with Planner and without Planner.

\end{quote}
\end{abstract}

\section{Introduction}
\noindent A lot of recent RL algorithms have proved to be very useful to solve simple tasks that were expected hard to solve by computers. These RL algorithms consist of an agent that learns a policy with the highest output reward. However, in a real-life scenario, we see the necessity of the interaction of multiple agents to ensure common goals are met. Yet, not a lot of progress has been seen concerning multiple agents. We show that the tasks are learned effectively with the help of multiple agents.

Secondly, to coordinate multiple agents, a planner is needed. The planner coordinates different tasks to different agents and ensures that policies are learned quickly. In this way, we see significant improvements in the training concerning the recent researches without a planner architecture.

Therefore, our contributions can be summarized as follows:
\begin{itemize}
    \item We coordinate learning of multiple agents with a central controller.
    \item We make the learning of the policies more efficient with the help of a planner, that assigns individual tasks to each agent.
\end{itemize}

\section{Related Work}
Coordinating multiple agents simultaneously is a challenging task. Some research works \cite{natarajan2010multi,shoham2003multi} have focused on solving the tasks with reinforcement learning using multiple agents. However, coordinating multiple agents is hard. Without a proper control system and a good state representation, it becomes very hard to obtain the optimal policy quickly, since the size of state space is significantly high. To address this issue, \cite{natarajan2008learning} introduce First Order Conditional Inference (FOCI) statements, which is of the form ``if $condition$ then $x1$ influences $x2$" where $condition$ and $x1$ are first-order literals.

To coordinate multiple agents, we also use a heuristic planner based on Manhattan Distances between source and target, similar to ~\cite{kokel2021reprel}. The central controller handles the planner and assigns available tasks to free agents, which ensures efficiency and obtaining optimal policy quickly. In~\cite{kokel2021reprel}, authors use FOCI dynamically, which enables them abstracted state space. 

To efficiently make the agents learn the tasks, many hierarchical learning frameworks like Options Framework~\cite{stolle2002learning} have been used. The main task is divided into a sequence of small tasks, and each small task is learned using an RL algorithm. In~\cite{illanes2020symbolic}, authors use a planner along with Options Framework to train their agent to learn and transfer tasks.

Finally, to obtain an optimal policy, many RL algorithms rely on TD-update with QLearning~\cite{hester2018deep}. For QLearning, authors optimize Q-Values of state-action pair. Instead of the Monte-Carlo update, using TD-update simplifies the optimization problem. In our work, we use this optimization to obtain an optimal RL policy for each option in the Options Framework.

\section{Approach}
\subsection{Environment}
For an agent, we define the Markov Decision Process (MDP) of the environment as follows:

\begin{equation}
    M = \{S, A, R, P\}
\end{equation}
where states $S$ are the set of state representations of agent, action space $A$ is the set of primitive actions an agent can take at every time step. The action space consists of 5 primitive actions $\{"up", "down", "left", "right", "no-op"\}$. $R$ is the reward function defined on states $S$ and actions $A$ and state transition function $P: S\times A \times S \rightarrow \{0,1\}$.

\subsection{Planner}
Our Multi-Agent RL Framework leverages a planner based on Manhattan Distances. In every step, each agent with an unallocated target calls the planner. The planner assigns the closest target to the agent based on the Manhattan Distance of the agent from unassigned targets. Therefore, our planner handles each agent's actions based on their targets.

\subsection{Handling Multiple Agents}
We propose using a central controller to manage the learning of all $n$ agents. To leverage the benefit of a central controller, we treat each agent independently. Each agent independently manipulates the Q-Tables of the central controller and takes action. To handle each agent individually, we need to abstract the state of the environment concerning an agent $i$ from $1$ to $n$. For an agent $i$, we define its state representation $\hat{s}_i$. We get this state representation $\hat{s}_i$ using the D-FOCI statements. Using this state representation, the agent $i$ obtains an action $a_i \in A$ according to the Q-Value of the central controller at that state. An example of D-FOCI statements can be given through the task of an agent $i$ at $l1$ picking its assigned gem $j$ at $l2$. Since the state agent, $i$ acquiring gem $j$ is not dependent on any other agent or gem, we define our D-FOCI statement only based on agent $i$ and gem $j$ as follows:
\begin{align*}
   & \{agent-at(i,l1), gem-at(j,l2), \\
   & assigned-agent-gem(i,j)\} \rightarrow acquired-gem(i, j)
\end{align*}

\subsection{MACOptions}
We define a central controller that optimizes the overall sum of the rewards obtained by each agent. The central controller of the environment $M$ tracks the action of all $n$ agents and assigns an action to each agent. We obtain the optimal policy for a central controller using the Options Framework. For each option $o$, all the agents update the value $q_{\pi_o}$ of a single Q-Table of the central controller, and exploit the values of Q-Table for the next action, using policy $\pi_o$. Finally, the central controller returns an action $a_i \in A$ to an agent $i$ having state representation $\hat{s}_i$ using optimal policy $\pi_o^*$ for the option.
\begin{equation}
    \pi_o^*(\hat{s}_i) \leftarrow \argmaxB_{a_i \in A} q^*_{\pi_o}(\hat{s}_i,a_i)
\end{equation}
Over a finite horizon, the value $q^*_{\pi_o}(\hat{s},a)$ for $\hat{s}$ as an agent's state representation and $a \in A$ as the agent's action is obtained as follows:
\begin{equation}
    q^*_{\pi_o}(\hat{s},a) = R(\hat{s},a) + \sum_{\hat{s}' \in S}P(\hat{s},a,\hat{s}')\max_{a' \in A}q_{\pi_o}(\hat{s}',a')
\end{equation}

\section{Experiments}

\begin{figure}
    \centering
    \includegraphics[width=0.99\linewidth]{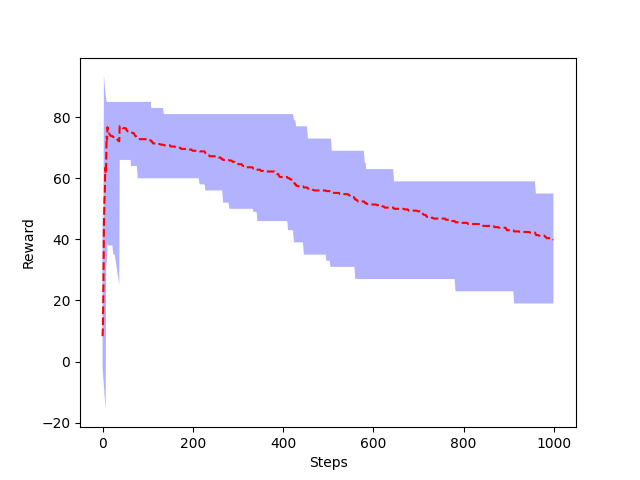} (a) QLearning
    \includegraphics[width=0.99\linewidth]{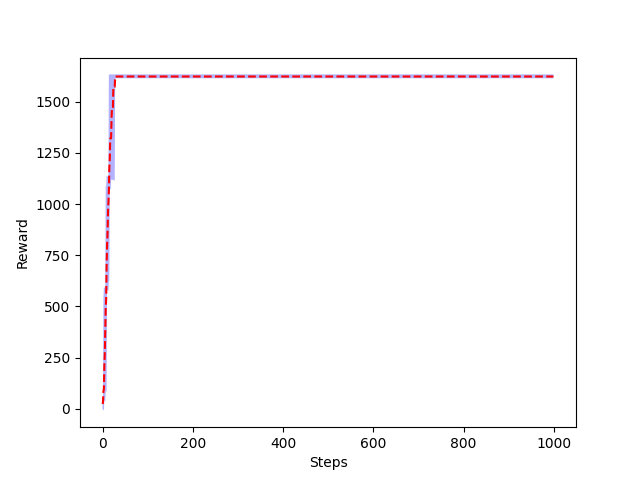} (b) QLearning + Options
    \includegraphics[width=0.99\linewidth]{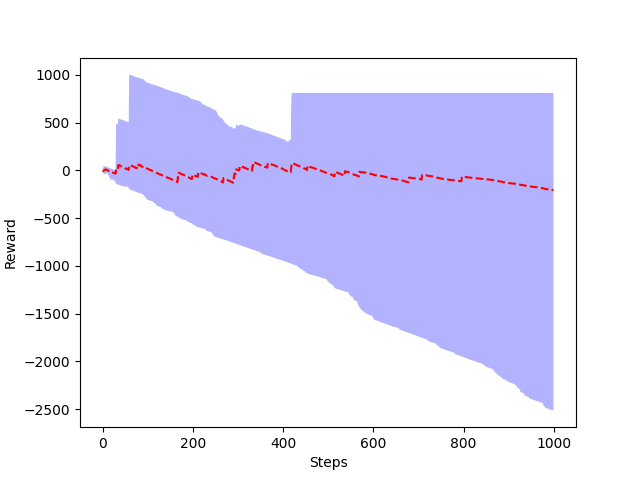} (c) Random Policy
    \caption{Test Plots}
    \label{fig:test}
\end{figure}

The experiments are run on a variation of box world environments which we call ``Bank world". Multiple agents are tasked with collecting gems and depositing them in a bank. The gems are located at different locations in a grid and the bank is located in the center of the grid. The agents aim to collect all the gems and deposit them all in the bank in the minimum possible steps. An agent can hold only one gem at a time. In our experiments, we use $2$ agents and $3$ gems.

\begin{figure}[ht]
    \centering
    \includegraphics[width=0.99\linewidth]{fig/test_plot_options.png} (a) With Planner
    \includegraphics[width=0.99\linewidth]{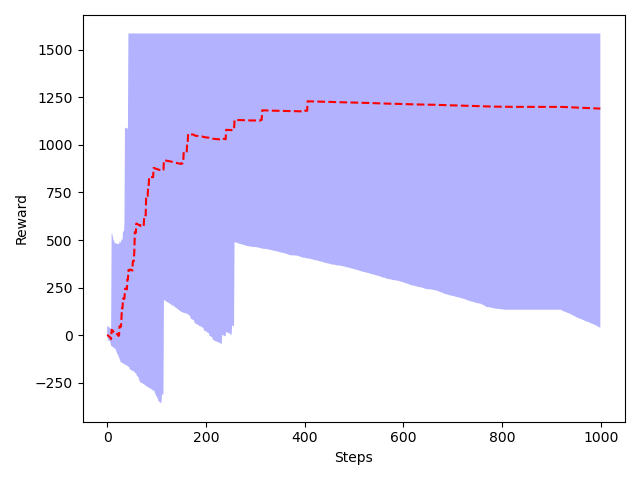} (b) Without Planner
    \caption{Effectiveness of Planner}
    \label{fig:test_planner}
\end{figure}

Our reward function $R$ is defined as follows:
\begin{equation}
    R(s,a) = \begin{cases}
    -5 & \text{$a$ is illegal. eg. hits the wall}\\
    50 & \text{allocated-gem is acquired by agent}\\
    500 & \text{allocated-gem is dropped at bank by agent}\\
    -1 & \text{otherwise}
    \end{cases}
\end{equation}

We conduct the experiments in 2 parts. In one set of experiments, we study the performance of different ways of training policy with the planner. In the second set of experiments, we study the importance of the planner, by comparing the performance of our method with the planner and without the planner. The experiments are explained in the following subsections.

\subsection{Obtaining Policies with Planner}
We keep a budget of 6k episodes for the training, where each episode has 1k steps.
Once the training is complete we run the experiment again with the learned Q values, 10 times and note the total reward achieved by the agents in 1k steps. Furthermore, in each experiment, the agents have access to the planner and we use the succinct state representation we get via the D-FOCI statements.

\begin{figure}
    \centering
    \includegraphics[width=0.99\linewidth]{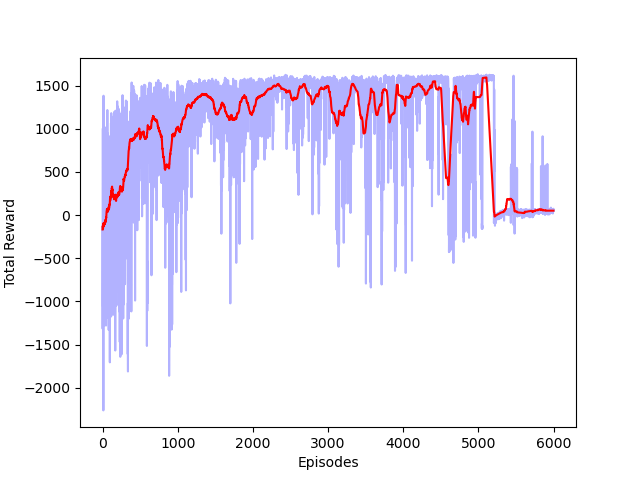} (a) QLearning
    \includegraphics[width=0.99\linewidth]{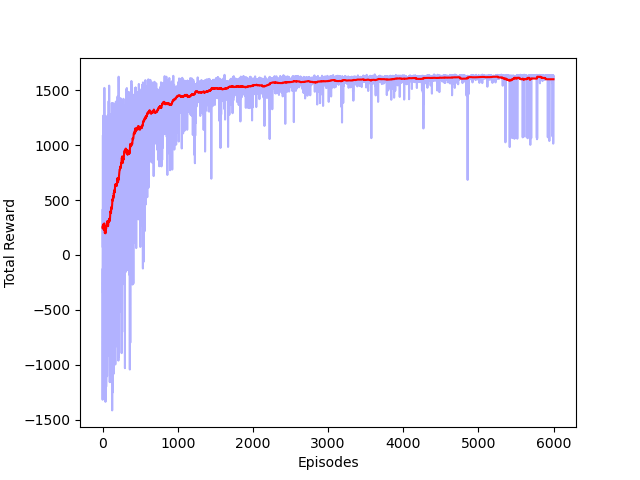} (b) QLearning + Options
    \includegraphics[width=0.99\linewidth]{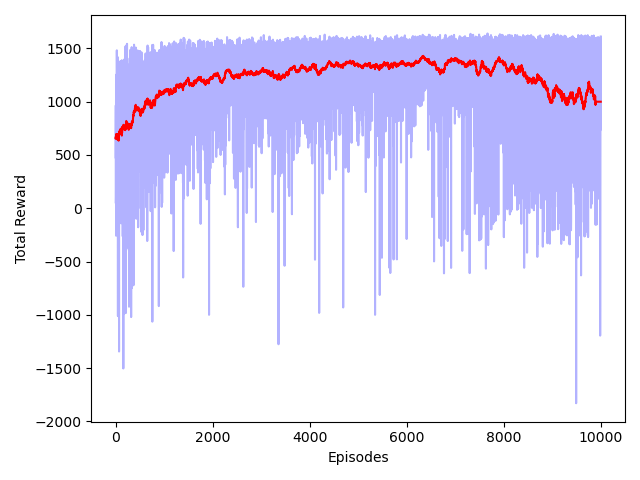} (c) QLearning + Options (no planner)
    \caption{Training Plots for (a) QLearning, (b) QLearning with Options and (c) QLearning + Options without Planner, showing total reward obtained per episode. }
    \label{fig:train}
\end{figure}

\noindent\textbf{Random Policy}: We run the experiment such that each agent follows a random policy.
Note, there is no learning in this case. This experiment is done just to establish a baseline so that comparisons can be made by learning a good policy in later experiments.

\noindent\textbf{Q-learning}: We define this experiment such that the central controller maintains a single Q-table which is updated using the TD update by each agent at every step. The idea is to learn the task as a whole by only looking at the rewards that the agents get at each step.

\noindent\textbf{Q-learning + Options}: We then define 2 options that learn 2 subtasks separately. One option learns the $pickup(A_i, G_i)$ task and the second option learns the $drop(A_i, G_j)$ task. Each option acts as an operator with its own D-FOCI statements as follows:\\

\noindent $\textbf{pickup-gem}(A_i, G_j)$:
\begin{align*}
    &  \{agent-at(i, l1), gem-at(j, l2), \\
    & assigned-agent-gem(i, j)\} \rightarrow acquired-gem(i, j)
\end{align*}

\noindent $\textbf{drop}(A_i, G_j)$:
\begin{align*}
    &  \{acquired-gem(i, j), agent-at(i, bank-loc)\} \rightarrow dropped(j)
\end{align*}

The central controller maintains 2 separate Q-tables for the sub-tasks. Each agent updates the respective Q-table for the option that it executes. We run the experiment finally with a similar setup as the other experiments.

\subsection{Performance with and without Planner}
We conduct the following experiments to determine the significance of using a planner. We compare the performance of the Q-learning + Options(with planner) model with Q-learning + options (without planner). The state representation, in this case, becomes a little different as there is no planner to tell the agents which gem to pick. Now each agent needs to know the location of all the gems 

\subsection{Results}
We observe in Fig.~\ref{fig:test} that the Q-learning + Options gives the maximum average reward in the test runs. The simple Q-learning gives the second-best average reward and the random policy gives the least. The results are exactly what we hypothesized. The use of options seems to have decreased the training time significantly. The comparison of training plots can be seen in Fig.~\ref{fig:train}. The simple Q-learning approach under the 6k episode budget doesn't seem to learn well enough although theoretically, it would have converged to the same results as the Q-learning with Options approach if there were a bigger episode budget. There is a significant difference between the average reward achieved by the random policy and that achieved by both of the other approaches which signify that the learning is significant.

In Fig.~\ref{fig:test_planner}, we see that using a planner makes our algorithm more target-oriented, therefore learning optimal policy quickly.

\section{Discussion and Future work}
One interesting observation that we made was if the number of agents is greater than the number of gems then the agents who do not get assigned to a gem just keep executing $"no-op"$ so that they don't gain a negative reward. It is impressive that such strategies can be learned.
Currently, the model is independent of the number of agents and gems but is dependent on the grid dimensions. Considering the bank will always be at the center of the grid we can train the agents by sampling random positions for the agent, gems, and also a random grid dimension at each episode. Further comparisons can be drawn by using a function approximator to predict the Q values instead of learning them via the bellman update. A future direction could be to see how the Q-learning + Options approach performs when the gems are also moving randomly at each step. 
Furthermore, this technique of using a centralized controller with a planner for multi-agents can be used in other domains to study its effectiveness in solving the task at hand.

\bibliography{ref.bib}

\end{document}